\documentclass[conference]{IEEEtran}
\usepackage{amsopn}
\usepackage{amsthm}
\usepackage{etex}
\usepackage{spconf,graphicx}
\usepackage{endnotes}
\usepackage{hyperref}
\usepackage{epsfig,psfrag}
\usepackage{pst-all}
\usepackage{amssymb,amsfonts,upref,cite,epsf,color,bm}
\usepackage{graphicx}
\usepackage{color}
\usepackage{amsmath}
\usepackage{graphicx}
\usepackage{calc}
\usepackage{booktabs}
\usepackage{tikz}
\usepackage{pgfplots}
\newcommand\defeq{:=}

\usepackage{algorithm}
\usepackage{algpseudocode}
\floatname{algorithm}{Algorithm}
\algnewcommand\algorithmicinput{\textbf{Input:}}
\algnewcommand\INPUT{\item[\algorithmicinput]}
\algnewcommand\algorithmicoutput{\textbf{Output:}}
\algnewcommand\OUTPUT{\item[\algorithmicoutput]}
\newtheorem{assumption}{Assumption}

\DeclareMathOperator*{\argmin}{arg\;min}

\newcommand\vect[1]{\mathbf #1}

\newcommand{\vx}{\vect{x}}  
  
\newcommand{\vz}{\vect{z}}

\newcommand{\mW}{\mathbf{W}}
\newcommand{\measlen}{M}

\newcommand{\signalsize}{N}
\newcommand{\noise}{\varepsilon}

\newcommand{\graphsigs}{\mathbb{R}^{\mathcal{V}}}

\newcommand{\edges}{\mathcal{E}}
\newcommand{\cluster}{\mathcal{C}}
\newcommand{\nodes}{\mathcal{V}}
\newcommand{\graph}{\mathcal{G}}
\newcommand{\samplingset}{\mathcal{M}}
\newcommand{\clusteredsig}{x_{c}}
\newcommand{\edgeset}{\mathcal{S}}
\newcommand{\sigdim}{1}
\newcommand{\flow}{h}
\newcommand{\xsig}{x[\cdot]}
\newcommand{\xsigval}[1]{x[{#1}]} 
\newcommand{\zsig}{z[\cdot]}
\newcommand{\zsigval}[1]{z[{#1}]} 
\newcommand{\estxsig}{\hat{x}[\cdot]}
\newcommand{\estxsigval}[1]{\hat{x}[{#1}]} 
\newcommand{\partition}{\mathcal{F}}
\newcommand{\clusteredsigs}{\mathcal{X}}

\newtheorem{theorem}{Theorem}
\newtheorem{definition}[theorem]{Definition}
\newtheorem{lemma}[theorem]{Lemma}

\usepackage{setspace}

\title{Recovery Conditions and Sampling Strategies for Network Lasso}
\name{Alexandru Mara and Alexander Jung}
\address{\normalsize Department of Computer Science, Aalto University, Finland; firstname.lastname(at)aalto.fi\\[-0.5mm]
}

\begin{document}
	\maketitle
\begin{abstract}
The network Lasso is a recently proposed convex optimization method for machine learning 
from massive network structured datasets, i.e., big data over networks. It is a variant of the well-known 
least absolute shrinkage and selection operator (Lasso), which is underlying many methods in learning 
and signal processing involving sparse models. Highly scalable implementations of the network Lasso 
can be obtained by state-of-the art proximal methods, e.g., the alternating direction method of multipliers (ADMM). 
By generalizing the concept of the compatibility condition put forward by van de Geer and B{\"u}hlmann as a 
powerful tool for the analysis of plain Lasso, 
we derive a sufficient condition, i.e., the network compatibility condition, on the underlying network topology 
such that network Lasso accurately learns a clustered underlying graph signal. 
This network compatibility condition, relates the the location of the sampled nodes with the clustering structure 
of the network. In particular, the NCC informs the choice of which nodes to sample, or in machine learning terms, 
which data points provide most information if labeled. 
 
\end{abstract}


\section{Introduction}
 \label{sec_intro}

We consider semi-supervised learning from massive heterogeneous datasets with an intrinsic network structure which occur 
in many important applications ranging from image processing to bioinformatics  \cite{SandrMoura2014}. 
By contrast to standard supervised learning methods, e.g., linear or logistic regression, which embed the data points 
into euclidean space \cite{BishopBook,hastie01statisticallearning}, we model the data points as nodes of a finite space whose discrete topology is represented by 
data graph $\graph=(\nodes,\edges,\mathbf{W})$ with the nodes $\nodes$ representing individual data points. 
Two nodes $i,j \in \nodes$ which represent similar data points are connected by an edge $\{i,j\} \in \edges$ whose 
strength is quantified by the positive weight $W_{i,j}$.

The goal of semi-supervised learning for network structured datasets is to learn an underlying hypothesis which 
maps each data point $i \in \nodes$ to a label $x[i]$, which can be a categorial or continuous variable. In some 
applications we have access to a small amount of initial label information in the form of (typically corrupted) samples 
$x[i]$ taken for all nodes $i \in \samplingset$ in a small sampling set $\samplingset$. In order 
to learn the complete label information, we rely on a smoothness hypothesis \cite{BishopBook,SemiSupervisedBook}, 
requiring the signal to be nearly constant over well connected subset of nodes (clusters). 

By representing label information as graph signals and using their total variation (TV) 
for measuring smoothness of the labeling, the learning problem can be 
formulated as a convex TV minimization problem. 
Following this approach, the authors of \cite{NetworkLasso} obtain the network 
Lasso which can be interpreted as a generalization of Lasso based method for 
learning sparse parameters \cite{hastie01statisticallearning}. 

An efficient scalable implementation of the network Lasso can be obtained via the 
alternating direction method of multipliers (ADMM) \cite{DistrOptStatistLearningADMM}. 
The implementation via ADMM is appealing since the resulting iterative algorithm is 
highly scalable, by using modern big data frameworks, and guaranteed to converge 
under the most general conditions \cite{DistrOptStatistLearningADMM}. 
 
In this paper, we present a condition on the network topology such that network Lasso is able 
to accurately learn a clustered graph signal. To this end, we introduce a very simple model 
for graph signals which are constant over a well connected group of nodes (clusters). Our condition, 
which we coin ``network compatibility condition'' amounts to the existence of certain network flows and 
is closely related to the ``network nullspace condition'' proposed recently by the first author \cite{RandomWalkSampling,NNSPSampta2017}.  

The closest to our research program, initiated by the works \cite{RandomWalkSampling,NNSPSampta2017,JungHero2016,JungSpawc2016,HannakAsilomar2016,Elsamlou2016}, 
is \cite{SharpnackJMLR2012,TrendGraph}, which provide sufficient conditions such that a special 
case of the network Lasso (referred to as the  ``edge Lasso'') accurately recovers smooth graph signals 
from noisy observations. However, these works require access to fully labeled 
datasets, while we consider datasets which are only partially labeled.

{\bf Outline.} We formalize the problem of recovering (learning) smooth graph signals 
from observing its values at few sampled nodes in Section \ref{sec_setup}. 
In particular, we show how to formulate this recovery as a convex optimization problem 
which coincides with the network Lasso problem studied in \cite{NetworkLasso}. 
Our main result, stated in Section \ref{sec_main_results}, is a sufficient condition on 
the network structure and sampling set such that accurate recovery is possible. 
Loosely speaking, this condition requires to sample nodes 
which are well-connected to the boundaries of clusters. 

\section{Problem Formulation}
\label{sec_setup}

\begin{figure}
\hspace*{0.5em}\includegraphics[width=1\columnwidth,angle=0]{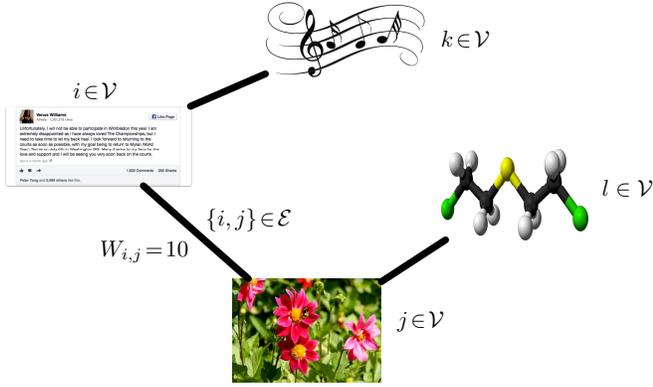}
\caption{\label{fig_hetero}A heterogeneous dataset represented by a data graph with individual data points as nodes.}
\end{figure}

We consider massive heterogeneous datasets which are represented 
by a network, i.e., a undirected weighted data graph $\graph\!=\!(\nodes,\edges,\mathbf{W})$ nodes $\nodes$ 
represent individual data points. For example, the node $i \!\in\! \nodes$ 
might represent a chat message on a user profile, measurements of a molecule, a sound fragment or 
a tabulated numerical data (cf.\ Figure \ref{fig_hetero}) \cite{BigDataNetworksBook}. 

Many applications naturally suggest a notion of similarity between individual 
data points, e.g., the profiles of befriended social network users or greyscale 
values of neighbouring image pixels. These domain-specific notions of similarity are represented by the 
edges of the graph $\graph$, i.e., the nodes $i,j\!\in\!\nodes$ representing similar 
data points are connected by an undirected edge $\{i,j\}\!\in\!\edges$. 
We quantify the extent of the similarity between connected data 
points $\{i,j\} \in \edges$ using positive edge weights $W_{i,j}\!>\!0$, which we 
collect in the symmetric weight matrix $\mW \in \mathbb{R}_{+}^{\signalsize \times \signalsize}$.  
In what follows, we consider only simple data graphs without self loops, i.e., for any $i \in \nodes$ we have 
$\{i,i\} \notin \edges$ and $W_{i,i}=0$. 

We sometimes need to orient the data graph $\graph=(\nodes,\edges,\mathbf{W})$ by declaring for 
each edge $e=\{i,j\} \in \edges$ one node as the head $e^{+}$ (e.g., $e^{+}=i$) and 
the other node as the tail $e^{-}$ (e.g., $e^{-}=j$) . Given an edge set $\edgeset$ in the 
data graph $\graph$, we denote the set of directed edges obtained by orienting $\graph$ 
as $\overrightarrow{\edgeset}$. 

Beside the network structure, encoded by the edges $\edges$, a dataset typically contains additional information, 
e.g., features, labels or model parameters associated with individual data points. 
Let us represent this additional information by a graph signal defined over the data graph $\graph$. 
A graph signal $\xsig$ is a mapping $\nodes \rightarrow \mathbb{R}^{\sigdim}$, which 
associates every node $i\!\in\!\nodes$ with the value $\xsigval{i} \!\in\! \mathbb{R}$. 
For the house prize example considered in \cite{NetworkLasso}, the graph signal $\xsigval{i}$ corresponds to 
a regression parameter for a local prize model (used for the house market in a limited geographical 
area represented by the node $i$). 
\begin{figure}
\hspace*{0.5em}\includegraphics[width=1\columnwidth,angle=0]{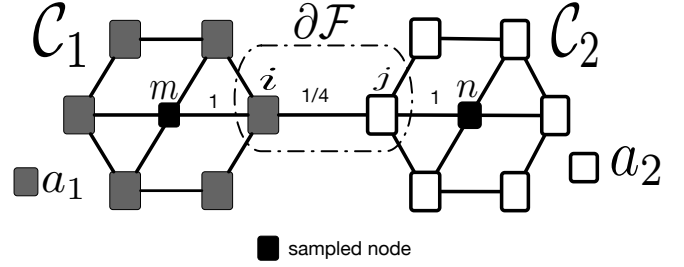}
\caption{\label{fig_graph_signals}Data graph $\graph=(\nodes,\edges,\mathbf{W})$ partitioned into two clusters $\partition=\{\cluster_{1},\cluster_{2}\}$ which 
are connected by the boundary edge $\{i,j\} \in \edges$. The underlying graph signal, which is constant over nodes from the same cluster, 
is sampled on the sampling set $\samplingset=\{m,n\}$.}
\label{fig_two_clusters}
\end{figure} 
In some applications, initial labels $y_{i}$ are available for few data points $i$ only. 
We collect those nodes $i\in \nodes$ in the data graph $\graph$ for which initial 
labels are available in the sampling set $\samplingset \subseteq \nodes$ (typically $|\samplingset| \ll |\nodes|$). 
In what follows, we model the initial labels as noisy versions of the true underlying labels $\vx[i]$, i.e., 
\begin{equation}
\label{equ_model_initial_labels}
y_{i} = \xsigval{i} + \noise[i]. 
\end{equation}  

\subsection{Learning Graph Signals}
\label{equ_gsr_sec}
We aim at learning a graph signal $\xsig \in \graphsigs$ defined over the 
date graph $\graph$, from observing its noisy values $\{ y_{i} \}_{i\!\in\!\samplingset}$ 
provided on a (small) sampling set
\begin{equation} 
\samplingset\defeq\{i_{1},\ldots,i_{\measlen}\} \subseteq \nodes,
\end{equation}
where typically $\measlen \ll \signalsize$. 

The network Lasso, is a particular recovery method which rests 
on a smoothness assumption, which is similar in spirit 
to the smoothness hypothesis of supervised machine 
learning \cite{SemiSupervisedBook}: 
\begin{assumption}
\label{ass_informal}
The graph signal values (labels) $\xsigval{i}$, $\xsigval{i}$ of two nodes $i,j \in \nodes$ 
within a cluster of the data graph $\graph$ are similar, i.e., $\xsigval{i} \approx \xsigval{j}$. 
\end{assumption} 
The class of smooth graph signals includes low-pass 
signals in digital signal processing where time samples at 
adjacent time instants (forming a chain graph) are strongly correlated for sufficiently 
high sampling rate. Another application involving smooth 
graph signals is image processing for natural images (forming a grid graph) whose 
close-by pixels tend to be coloured likely. 

What sets our work apart from digital signal processing, is that we consider datasets 
whose data graph is not restricted to regular chain or grid graphs but may form an 
arbitrary (complex) networks. In particular, our analysis targets the tendency of the networks 
occurring in many practical applications to form clusters, i.e., well-connected subset of nodes. 
A very basic example of such a clustered data graph is illustrated in Figure \ref{fig_graph_signals}, which involves 
a partition of the data graph into two disjoint clusters $\cluster_{1}$ and $\cluster_{2}$. 
The informal smoothness hypothesis Assumption \ref{ass_informal} required the signal valued $x[i]$ for 
all nodes $i \in \cluster_{1}$ (or $i\in \cluster_{2}$) to be mutually similar, e.g., to the value $a_{1}$ (or $a_{2}$). 

In what follows, we will quantify the smoothness of a graph signal $\xsig \in \graphsigs$ 
via its \emph{total variation} (TV)
\begin{equation} 
\label{equ_def_TV}
\| \xsig \|_{\rm TV} \defeq \sum_{\{i,j\} \in \edges} W_{i,j}  | \xsigval{j} \!-\! \xsigval{i}| . 
\end{equation} 
It will be convenient to introduce, for a given subset of edges $\edgeset \subseteq \edges$, 
the shorthand 
\begin{equation}
\| \xsig \|_{\edgeset} \defeq \sum_{\{i,j\} \in \edgeset} W_{i,j}  | \xsigval{j} \!-\! \xsigval{i}| . 
\end{equation}
Besides smoothness another criterion for learning graph signals $\estxsig$ 
is a small empirical error 
\begin{equation}
\label{equ_def_emp_error}
\widehat{E}(\estxsig) \defeq \sum_{i \in \samplingset} | \estxsigval{i} - y_{i}|, 
\end{equation}
where $y_{i}$ denotes initial labels provided for all data points $i \in \samplingset$ 
belonging to the sampling set $\samplingset$.

Learning a signal with small TV $\| \estxsig \|_{\rm TV}$ and 
small empirical error $\widehat{E}(\estxsig)$ (cf.\ \eqref{equ_def_emp_error}), yields the optimization problem
\begin{align} 
\estxsig & \in \argmin_{\tilde{x}[\cdot] \in \graphsigs} \widehat{E}(\tilde{x}[\cdot])  + \lambda \| \tilde{x}[\cdot] \|_{\rm TV}.  \label{equ_semi_sup_learning_problem}
\end{align}
As the notation already indicates, there might be multiple solutions $\estxsig$ 
for the optimization problem \eqref{equ_semi_sup_learning_problem}. However, 
any learned graph signal $\estxsig$ obtained by solving \eqref{equ_semi_sup_learning_problem} 
balances the empirical error $\widehat{E}(\estxsig)$ with the TV $ \| \estxsig \|_{\rm TV}$ 
of the learned graph signal. The optimization problem \eqref{equ_semi_sup_learning_problem} is a 
special case of the network Lasso problem studied in \cite{NetworkLasso}. In particular, the network 
Lasso formulation in \cite{NetworkLasso} allows for vector valued labels $\vx[i] \in \mathbb{R}^{p}$ and more general 
empirical loss functions. The parameter $\lambda$ in \eqref{equ_semi_sup_learning_problem} 
allows to trade off small empirical error against signal smoothness. In particular, choosing a 
small value for $\lambda$ enforces the solutions of \eqref{equ_semi_sup_learning_problem} 
to yield a small empirical error, whereas choosing a large value for $\lambda$ enforces the 
solutions of \eqref{equ_semi_sup_learning_problem} to have small TV, i.e., to be smooth. 

There exist highly efficient methods for solving the network Lasso problem 
\eqref{equ_semi_sup_learning_problem} (cf.\ \cite{ZhuAugADMM} and the references therein). 
Most of the state-of-the art convex optimization method belong to the family of 
proximal methods \cite{ProximalMethods}. One particular instance of proximal methods is 
ADMM which has been applied to the network Lasso in \cite{NetworkLasso} to obtain 
a highly scalable learning algorithm.

\section{Network Compatibility Condition} 
\label{sec_main_results} 

For network Lasso methods, based on solving \eqref{equ_semi_sup_learning_problem}, to be accurate,  
we have to verify the solutions $\estxsig$ of 
\eqref{equ_semi_sup_learning_problem} to be close to the true (but unknown) underlying graph signal $\xsig \in \graphsigs$. 
In what follows, we present a condition which guarantees any 
solution $\estxsig$ of \eqref{equ_semi_sup_learning_problem} to be close to a clustered graph signal $\xsig$. 
Given a fixed partition $\partition=\{\cluster_{1},\cluster_{2},\ldots\}$ of the data graph into disjoint clusters 
$\cluster_{l} \subset \nodes$, we define the class of \emph{clustered graph signals} by 
\begin{equation}
\label{equ_def_clustered_signal_model}
\xsig\!\in\! \clusteredsigs \!\defeq\! \{ \clusteredsig[\cdot] \!\in\! \graphsigs:   \clusteredsig[i] \!=\!\sum_{\cluster \in \partition} a_{\cluster} \mathcal{I}_{\cluster}[i] \} , 
\end{equation} 
where, for a subset $\cluster \subseteq \nodes$, we define the indicator signal 
\begin{equation}
\mathcal{I}_{\cluster}[i] \!\defeq\! \begin{cases} 1 &\mbox{ if } i \in \cluster \\ 0 & \mbox{ else.} \end{cases} 
\end{equation} 
For a given partition $\partition=\{\cluster_{1},\cluster_{2},\ldots\}$, the boundary 
$\partial \partition \subseteq \edges$ is the set of 
edges $\{i,j\} \in \edges$ which connect nodes $i \!\in\! \mathcal{C}_{a}$ and 
$j \!\in\! \mathcal{C}_{b}$ from different clusters, i.e., with $\mathcal{C}_{a} \!\neq\! \mathcal{C}_{b}$. 
For a partition $\partition = \{ \cluster_{1},\ldots,\cluster_{|\mathcal{F}|} \}$ whose overall 
boundary weight $\sum_{\{i,j\} \in  \partial \partition} W_{i,j}$ is small, 
the clustered graph signals \eqref{equ_def_clustered_signal_model} have small TV $\| \xsig \|_{\rm TV}$, i.e., 
they are smooth. 

The signal model \eqref{equ_def_clustered_signal_model}, 
which has been used also in \cite{SharpnackJMLR2012,TrendGraph}, 
is closely related to the stochastic 
block model (SBM) \cite{Mossel2012}. Indeed, the SBM is obtained from 
\eqref{equ_def_clustered_signal_model} by choosing the coefficients $a_{\cluster}$ uniquely
for each cluster, i.e., $a_{\cluster} \in \{1,\ldots,|\mathcal{F}|\}$. 
Moreover, the SBM provides a generative (stochastic) model 
for the edges within and between  the clusters $\cluster_{l}$. 


The main contribution of this paper is the insight that network Lasso accurately learns 
clustered graph signals (cf.\ \eqref{equ_def_clustered_signal_model}) if there exist certain network flows \cite{KleinbergTardos2006} 
between the sampled nodes in $\samplingset$. 
\begin{definition} 
Consider an empirical graph $\graph$ with an arbitrary but 
fixed orientation. A flow with demands $d[i] \in \mathbb{R}$, for $i \in \nodes$, 
is a mapping $\flow[\cdot]: \overrightarrow{\edges} \rightarrow \mathbb{R}_{+}$ satisfying 
\begin{itemize} 
\item the conservation law 
\begin{equation} 
\hspace*{-6mm}\sum_{j \in \mathcal{N}^{+}(i)}\hspace*{-2mm} \flow[(j,i)] \!-\! \hspace*{-2mm}\sum_{j \in \mathcal{N}^{-}(i)} \hspace*{-2mm}\flow[(i,j)] = d[i] \mbox{, for any }  i \!\in\! \nodes
\end{equation}
\item 
the capacity constraints 
\begin{equation} 
\flow[e] \leq W_{e} \mbox{ for any oriented edge } e \!\in\!  \overrightarrow{\edges}. 
\end{equation} 
\end{itemize}
\end{definition} 
Here, we used the directed neighbourhoods 
$\mathcal{N}^{+}(i)\!\defeq\!\{ j\!\in\!\mathcal{N}(i)\!:\!\{i,j\}^{+}\!=\!\{i\} \}$ and 
$\mathcal{N}^{-}(i) \!\defeq\!\{ j\!\in\!\mathcal{N}(i)\!:\!\{i,j\}^{-}\!=\!\{i\} \}$.

Using the notion of a network flow with demands, we now adapt the 
compatibility condition introduced for learning sparse vectors with the Lasso \cite{GeerBuhlConditions} 
to learning clustered graph signals (cf.\ \eqref{equ_def_clustered_signal_model}) with the network Lasso \eqref{equ_semi_sup_learning_problem}. 
\begin{definition}
\label{def_sampling_set_resolves}
Consider a sampling set $\samplingset$ and a partition $\partition$ 
of the data graph $\graph$ into disjoint subsets. Then, the network compatibility 
condition (NCC) with parameters $K,L >0$ is satisfied by $\samplingset$ and $\partition$, if for any orientation of the 
edges in the boundaries $\partial \partition$, we can orient the remaining edges in 
$\edges \setminus \partial \partition$ such that there exists a flow $\flow[e]$ with demands 
$d[i]$ on $\overrightarrow{\graph}$ such that 
\begin{itemize} 
\item for any boundary edge $e\!\in\! \partial \partition$: $h[e]= L \cdot W_{e}$, 
\item for every sampled node $i \!\in\! \samplingset$: $|d[i]| \!\leq\! K$, 
\item for ever other node $i\!\in\!\nodes \setminus \samplingset$: $d[i]\!=\!0$.
\end{itemize} 
\end{definition} 

We are now in the position to state our main result, i.e., if a sampling set satisfies the NCC, 
then any solution $\estxsig$ of the network Lasso is an accurate proxy for a true underlying 
clustered graph signal $\xsig \in \clusteredsigs$ (cf.\ \eqref{equ_def_clustered_signal_model}).
\begin{theorem} 
\label{main_thm_exact_sparse}
Consider a clustered labeling $\clusteredsig[\cdot]\!\in\!\clusteredsigs$ (cf.\ \eqref{equ_def_clustered_signal_model})
and its noisy versions $y_{i}$ for samples nodes $\samplingset \subseteq \nodes$. 
If $\samplingset$ resolves the partition $\partition$ with parameters $K,L >0$,  
then any solution $\estxsig$ of the network Lasso \eqref{equ_semi_sup_learning_problem} with $\lambda \!\defeq\! 1/K$ satisfies 
\begin{equation}
 \| \estxsig \!-\!\xsig \|_{\rm TV}\!\leq\! (K\!+\!4/(L\!-\!1))  \sum_{i \in \samplingset} |\noise[i]|. 
\end{equation} 
\end{theorem} 

It is important to realize that the network Lasso problem \eqref{equ_semi_sup_learning_problem} does not 
require knowledge of the partition $\partition$ underlying the unknown clustered graph signal $\vx[\cdot] \in \clusteredsigs$. 
The partition is only used for the analysis of learning methods based on the network 
Lasso \eqref{equ_semi_sup_learning_problem}. Moreover, for graph signals $\vx [\cdot]$ having different signal values over different 
clusters, the solutions of \eqref{equ_semi_sup_learning_problem} could be used for 
determining the clusters $\cluster_{k}$ which constitute the partitioning $\partition$. 

Finally, we point to the fact that the NCC depends on both: the sampling set $\samplingset$ and 
the graph partition $\partition$ via the (total weight of) boundary $\partial \partition$.  
Thus, for a given partition $\partition$, we might choose the sampling set 
$\samplingset$ such that the NCC is guaranteed. 
One particular such choice is suggested by the following result. 
\begin{lemma}
\label{lem_suff_cond_NNSP}
Consider a partitioning $\partition$ of the data graph $\graph$ which also contains the 
sampling set $\samplingset \subset \nodes$. If each boundary edge $\{i,j\} \in \partial \partition$ 
with  $i\!\in\!\mathcal{C}_{a}$, $j \!\in\! \mathcal{C}_{b}$ is connected to sampled nodes, i.e., 
$\{m,i\}\!\in\!\edges$ and $\{n,j\}\!\in\!\edges$ with $m \!\in\! \samplingset\!\cap\!\mathcal{C}_{a}$,  
$n\!\in\!\samplingset\!\cap\!\mathcal{C}_{b}$, and weights $W_{m,i}, W_{n,j} \geq L W_{i,j}$, 
then the sampling set $\samplingset$ satisfied NCC with parameters $K=L \max_{\{i,j\} \in \partition} W_{i,j}$ and $L$. 
\end{lemma} 
An application of Lemma \ref{lem_suff_cond_NNSP} to the data graph shown 
in Figure \ref{fig_two_clusters}, verifies the NCC of the sampling set $\samplingset=\{m,n\}$ 
with parameters $K=L=4$. 

Thus, according to Lemma \ref{lem_suff_cond_NNSP}, one should sample nodes which are close 
to the boundary between different clusters. There are highly scalable network algorithms available which 
aim at locating the boundaries of clusters \cite{Spielman2012}.

%

\section{Numerical Experiments}
\label{sec_numerical}

\begin{figure}
\begin{minipage}{.4\textwidth}
  \centering
  \hspace*{0em}\includegraphics[width=1\linewidth]{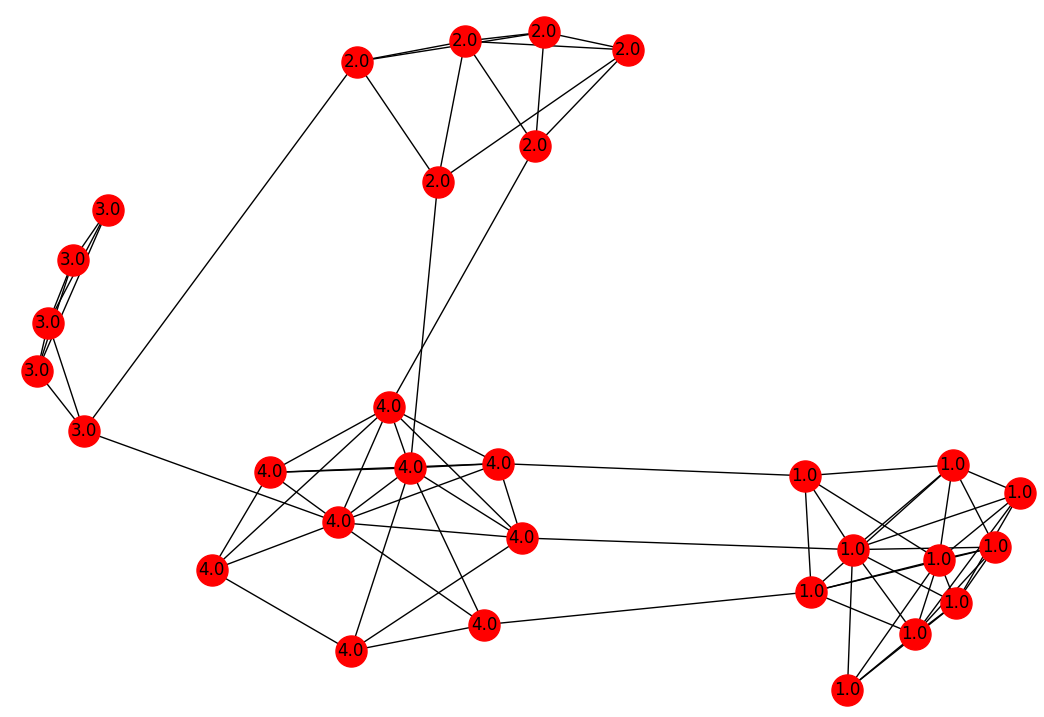}
  \caption{Data graph $\graph_{0}$ with signal values $x[i]$ indicated for each node.}
  \label{fig_signal}
\end{minipage}
\end{figure}

In order to illustrate the theoretical findings of Section \ref{sec_main_results} 
we applied network Lasso to a synthetically generated dataset with data graphs $\graph_{0}$. 
In particular, we generated a data graph $\graph_{0}$ using the popular LFR model proposed by Lancichinetti and Fortunato \cite{BenchmarkComDet}. 
The LFR model allows to generate networks with a community structure similar to  
to those of observed in many real-world networks. In particular, networks obtained from the LFR exhibit 
a power law distribution of node degrees and community sizes. 
The final synthetic data graph $\graph_{0}$ contains a total of $|\nodes| = 30$ nodes which are partitioned 
into four clusters $\partition=\{\cluster_{1},\cluster_{2},\cluster_{3},\cluster_{4}\}$. The nodes $\nodes$
are connected by $|\edges| = 156$ undirected edges $\edges$ with uniform edge weights $W_{i,j} = 1$ 
for all $\{i,j\} \in \edges$. Given the data graph $\graph_{0}$ and partition $\partition$ we generate 
a clustered graph signal according to \eqref{equ_def_clustered_signal_model} as 
$x_{0}[i] = \sum_{j=1}^{4} a_{j} \mathcal{I}_{\cluster_{j}}[i]$ with coefficients $a_{j}=j$. 
We illustrate the data graph $\graph_{0}$ along with the graph signal values $x_{0}[i]$ in Figure \ref{fig_signal}.

According to Lemma \ref{lem_suff_cond_NNSP}, in order to recover the entire graph signal $\xsig$ 
it is most helpful to have its signal values $\xsigval{i}$ for the nodes $i\in \nodes$ close to boundary $\partial \partition$ 
between different clusters. In order to verify this intuition, we constructed two different sampling sets. 
The first sampling set $\samplingset_{1}$ was constructed in line with Lemma \ref{lem_suff_cond_NNSP}, by prefering 
to sample nodes near a cluster boundary. By contrast, the second sampling set $\samplingset_{2}$ was obtained 
by selecting nodes uniformly at random and thus ignoring the cluster structure inherent to $\graph_{0}$. The size 
of both sampling sets is equal to $(1/2) |\nodes|$. For simplcity, we assume noiseless measurements, i..e, 
for the initial labels $y_{i}$ are given by $y_{i} = x[i]$ for each sampled node $i\!\in\!\samplingset$ (with either 
$\samplingset\!=\!samplingset_{1}$ or $\samplingset\!=\!\samplingset_{2}$).

\begin{figure}
\centering
\begin{minipage}{.5\textwidth}
  \centering
  \hspace*{0em}\includegraphics[width=1\linewidth]{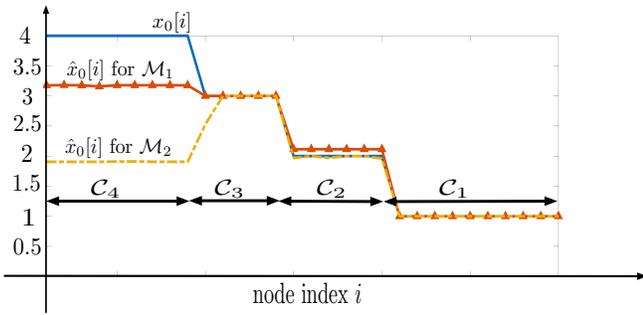}
  \caption{True graph signal along with the results of network Lasso when using sampling set $\mathcal{M}_{1}$ (Lemma \ref{lem_suff_cond_NNSP}) or $\mathcal{M}_{2}$ (random).}
  \label{fig_results}
\end{minipage}%
\end{figure}

For each of the two sampling sets $\samplingset_{1}$ and $\samplingset_{2}$, we learned the overall 
graph signal by solving the network Lasso problem \eqref{equ_semi_sup_learning_problem} using a modified version of the 
ADMM implementation discussed in \cite{NetworkLasso} (which considered the mean squared error instead 
of the empirical error \eqref{equ_def_emp_error}). The learned signals $\estxsig$ obtained finally for each of 
the two sampling sets are shown in Figure \ref{fig_results}, along with the true clustered graph signal $\xsig$. 


\section{Conclusions}
\label{sec5_conclusion}
We presented a sufficient condition on the network topology and sampling set such that 
any solution of the network Lasso problem is an accurate estimate for a true 
underlying clustered graph signal. This recovery condition, which we term the network 
compatibility condition, amounts to ensuring the existence of certain network flows 
with prescribed demands. We also provide a more specific, somewhat more practical, 
condition on the sampling set which implies the network compatibility condition. 
Loosely speaking, for a given budget of how man nodes to sample, our conditions 
suggest to sample more densely near to the boundaries between different clusters in 
the data graph. This intuition is verified by means of numerical experiments involving 
a toy data graph which has been generated in line with the properties of many real-world 
networks, i.e., presence of clusters and power-law degree distribution. 

\bibliographystyle{abbrv}
\bibliography{SLPbib}
\end{document}